\documentclass[11pt,a4paper]{article}
\usepackage[dvipsnames]{xcolor}
\usepackage{emnlp2018}
\usepackage{times}
\usepackage{latexsym}
\usepackage{lingmacros}
\usepackage{tcolorbox}

\usepackage{graphicx}
\usepackage{framed}
\usepackage{pifont}

\aclfinalcopy 

\title{Talking to myself: self-dialogues as data for conversational agents}

\author{Joachim Fainberg \ \ \  Ben Krause \ \ \  Mihai Dobre \ \ \  Marco Damonte \\ \ \ \  \textbf{Emmanuel Kahembwe} \ \ \  \textbf{Daniel Duma} \ \ \  \textbf{Bonnie Webber} \ \ \  \textbf{Federico Fancellu} \\
  School of Informatics\\
  University of Edinburgh  \\
  Edinburgh, UK \\
  {\tt \{j.fainberg,ben.krause,bonnie.webber,f.fancellu\}@ed.ac.uk}
  \\}

\date{}

\begin{document}
\maketitle
\begin{abstract}
Conversational agents are gaining popularity with the increasing ubiquity of smart devices. However, training agents in a data driven manner is challenging due to a lack of suitable corpora. This paper presents a novel method for gathering topical, unstructured conversational data in an efficient way: self-dialogues through crowd-sourcing. Alongside this paper, we include a corpus of 3.6 million words across 23 topics. We argue the utility of the corpus by comparing self-dialogues with standard two-party conversations as well as data from other corpora.
\end{abstract}

\section{Introduction}

Open-domain conversational agents have recently gained significant attention from the machine learning community through competitions such as the Amazon Alexa Prize\footnote{\url{developer.amazon.com/alexaprize}}. They have stoked public interest via devices such as the Amazon Echo and Google Home. By design, open-domain conversational agents require the ability to converse about a broad set of topics in a fluid and unconstrained manner while keeping dialogue with the end-user coherent, clear and engaging.

These requirements make it difficult and often impractical to use data-driven methods since the available conversational corpora are either too small, artificial, narrowly focused and often do not model any of the domain or the entities that a user may wish to talk about. Conversational corpora are also expensive to gather and require two people who have never met each other and may not have the same knowledge of a topic to hold a conversation.

This paper proposes a novel way to gather domain specific, conversational data in an efficient, cost-saving way: \textit{self-dialogues through crowd-sourcing}. Instead of a standard, two party conversation, self-dialogues are fictitious conversations orchestrated by one person who plays both parts in a dialogue. Using this technique we collect a corpus of approximately 3 million words across 23 topics via Amazon Mechanical Turk, that we make available alongside this paper\footnote{\url{github.com/jfainberg/self\_dialogue\_corpus}}. We initially collected and used this corpus in our entry to the Amazon Alexa Prize 2017 \cite{krause2017edina}, and present more detailed analysis of the dataset here.

We report two preliminary analyses of the corpus. First, we compare our dataset with standard two-party conversations and find that our setup not only halves costs, since one person is paid per conversation instead of two, but also leads to better coherence and engagement. We note here that the intended purpose of the dataset is to train conversational agents. Our discussions are limited to this scope and we make no claims about its linguistic properties.

We build a simple retrieval agent to return a response given a user query and compare the responses returned by our dataset against those returned by the OpenSubtitles corpus \cite{tiedemann2012parallel}. By qualitatively inspecting both general and topical queries, we find that our corpus is more suitable for open-domain human-bot conversational settings.
 
\section{Related corpora}
In this Section we discuss some related, publicly available work, but refer to \citet{serban2015survey} for a survey of existing dialogue corpora. However, we note that many of the corpora discussed therein are not publicly available\footnote{The Dialogue Diversity Corpus lists a range of available task-oriented corpora: \url{www-bcf.usc.edu/~billmann/diversity/DDivers-site.htm}}.

The Switchboard \cite{godfrey1992switchboard} corpus consists of roughly 3 million words of transcribed two-speaker phone conversations on 70 predefined topics. It is similar in the number of words to the work presented here, but consists of longer dialogues spread across more topics. The Ubuntu dialogue corpus \cite{W15-4640} is a large corpus of unstructured dialogues with approximately 100 million words. This corpus is collected from Ubuntu support chat logs and as such may have limited utility outside of that domain. The Dialog State Tracking Challenge (DSTC) \cite{williams2013dialog} aims to track what a user wants from an agent at each turn in a dialogue. The associated dataset is gathered using a slot-filling approach with users conversing with an existing machine.

Some corpora, such as OpenSubtitles \cite{tiedemann2012parallel} and Movie-DiC \cite{banchs2012movie}, gather conversational data from movie subtitles or scripts. While these corpora tend to be very large, they can be rather noisy, with multiple parties in conversations and occasionally the same party taking consecutive turns.

It is also possible to gather large amounts of unstructured data through services such as Reddit\footnote{\url{files.pushshift.io/reddit/comments}} or Twitter  \cite{ritter2010unsupervised,shang2015neural}. These data streams have the advantage of being domain-specific (by selecting specific sub-reddits or hash tags). However, their properties may vary significantly in terms of the number of parties in a conversation, the response length, and the quality of language. 

Finally, the VisDial dataset \cite{DBLP:journals/corr/DasKGSYMPB16} contains dialogues between two humans discussing a particular image. Their task is closely related to the work presented here, in that the dataset is collected through crowd sourcing. However, when adapting the task to fit our goal we found difficulties in managing the data collection. This is further discussed in Section~\ref{sec:corpus_discussion}.

\section{Self-dialogue Corpus}

This corpus was collected using Amazon Mechanical Turk (AMT). To harvest \textit{self-dialogues}, we asked Workers to create a fictitious two-party conversation around a topic.
For the majority of the tasks, the Workers were requested to fill 20 text boxes with a conversation on a particular topic. The setup required all 20 text boxes to be completed in order to submit.

\begin{figure}[htbp]

  \centering
  \fbox{\includegraphics[width=1.0\columnwidth]{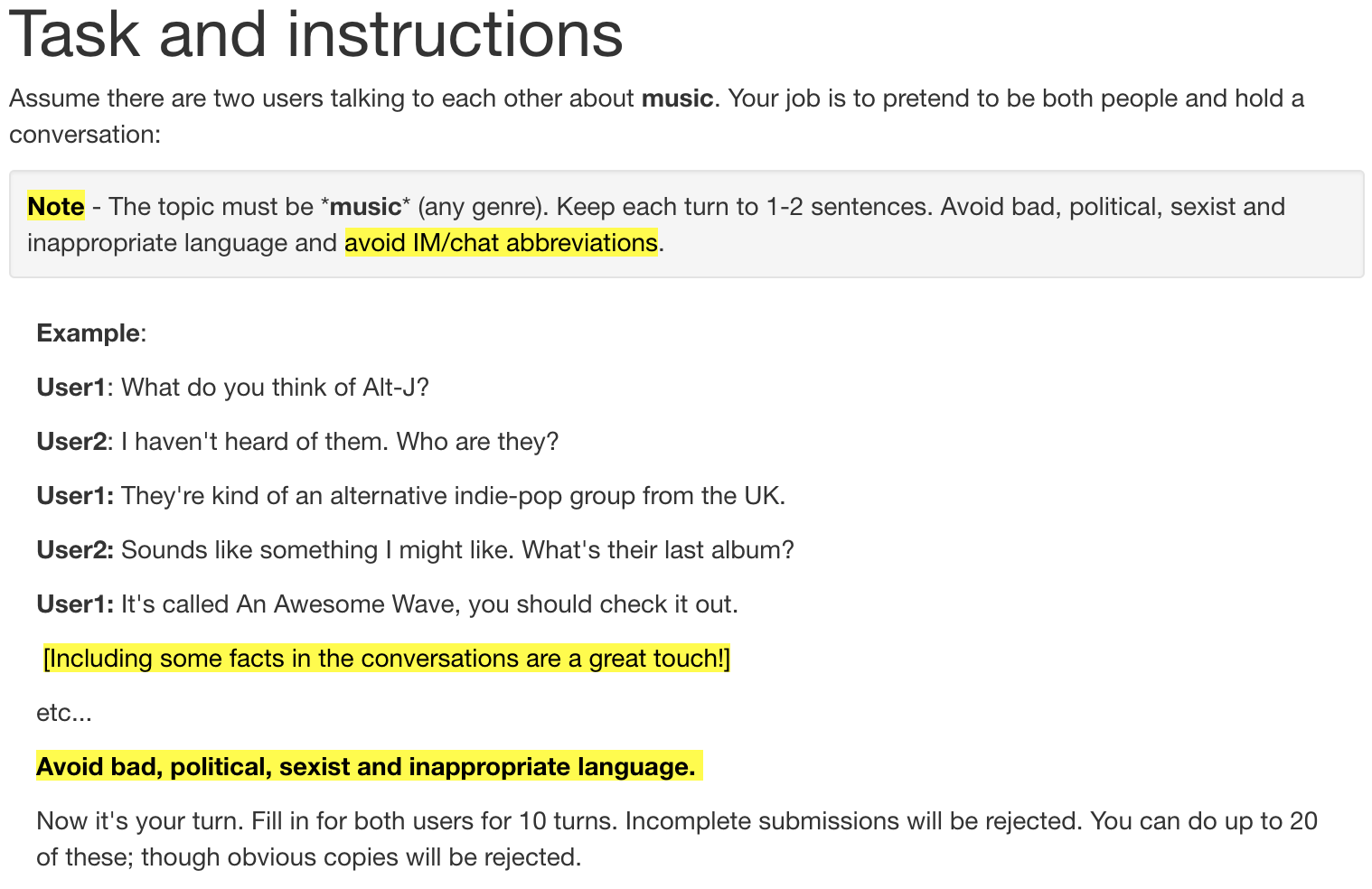}}
  \caption{Amazon Mechanical Turk instructions for the general music task. Other tasks followed a similar pattern with dedicated examples.}
  \label{fig:mturk_interface_self}
\end{figure}

In order to obtain conversations that were as natural as possible, we limited the number of constraints set in the task descriptions. We aimed to present tasks that were simple to execute. The only rejection criteria related to abusing the system, such as submitting (near) duplicate entries or content with exaggerated bad language. To deal with the large amount of submissions, we automated the rejection procedure by 1) comparing the cosine similarity between bags of words of two dialogues and 2) flagging conversations which contained a large number of words from a bad-words list. There is no monetary loss to rejecting a submission, but wrongful rejection may lead to poor reviews as a Requester and consequently slow down future tasks. In total only eight out of 2,717 Workers were banned, and 145 conversations ($\approx 0.6\%$) were rejected. An example of the interface shown to Workers is shown in Figure~\ref{fig:mturk_interface_self}.

After experimenting with pay, region requirements, and AMT credentials, we converged upon the following criteria which holds for the majority of the Workers in the corpus:
\begin{itemize}
\itemsep0em
\item location: United States (and territories), United Kingdom;
\item HIT approval rate: greater than 95\%;
\item number of HITs approved: greater than 500;
\item number of conversations per worker per batch: maximum 20;
\item pay per 10-turn conversation: US \$0.70-0.80; 5-turn conversation: US \$0.35-0.40.
\end{itemize}

A summary of the statistics of the dataset is shown in Table~\ref{tab:stats} and counts per topic are shown in Table~\ref{tab:data}.

\begin{table}[h]
\begin{center}
  \begin{small}
  \begin{tabular}{|l|l|}
    \hline
    \bf Category & \bf Count \\
    \hline
    Topics & 23 \\
    Conversations & 24,283\\
    Words & 3,653,313 \\
    Turns & 141,945 \\
    Unique users & 2,717 \\
    Avg. convos per user & $\sim$9 \\
    Peak convos per day & 2,307\\
    Unique tokens & 117,068 \\
    \hline
  \end{tabular}
  \end{small}
  \smallskip
    \caption{Summary statistics.}
    \label{tab:stats}
\end{center}
\end{table}



\begin{table}[htbp]
\begin{center}
  \begin{small}
  \begin{tabular}{|l|lll|}
    \hline
    \bf Topic/subtopic & \bf \# Conv. & \bf \# Words & \bf \# Turns \\
    \hline
    Movies & 4,126 & 814,842 & 82,018 \\
    Action & 414 & 37,037 & 4,140 \\
    Comedy & 414 & 36,401 & 4,140 \\
    Fast \& Furious & 343 & 33,964 & 3,430 \\
    Harry Potter & 414 & 44,220 & 4,140 \\
    Disney & 2,331 & 232,573 & 23,287 \\
    Horror & 414 & 428,33 & 4,138 \\
    Thriller & 828 & 77,975 & 8,277\\
    Star Wars & 1,726 & 178,351 & 17,260 \\
    Superhero & 414 & 40,967 & 4,140 \\\hline
    Music & 4,911 & 924,993 & 98,123 \\
    Pop & 684 & 62,383 & 6,840\\
    Rap / Hip-Hop & 684 & 66,376 & 6,840\\
    Rock & 684 & 63,349 & 6,837\\
    The Beatles & 679 &  68,396 & 6,781 \\
    Lady Gaga & 558 & 49,313 & 5,566\\\hline
    Music and Movies & 216 & 37,303 & 4,320 \\\hline
    Transition Mus-Mov & 198 & 4,287 & 396 \\\hline
    Baseball & 496 &  99,298 & 9,881 \\
    Basketball  & 485 & 95,264 & 9,507\\
    Ice Hockey  & 174 & 34,288  & 3,417 \\
    NFL Football & 2,801 &  562,801 & 55,939\\\hline
    Fashion & 210 & 46,099 & 4,105 \\
    \hline
  \end{tabular}
  \end{small}
  \smallskip
    \caption{Data collected divided by topic/tasks. `Transition Mus-Mov' refers to a short transition task, where Workers were asked to transition from any topic to either music or movies within one turn.}
    \label{tab:data}
\end{center}
\end{table}

The following example shows a self-dialogue in the `Disney Movies' category\footnote{See Appendix~A for more examples.}.

\begin{tcolorbox}[colback=white]
What is your favorite movie?\\
I think Beauty and the Beast is my favorite.\\
The new one?\\
No, the cartoon. Something about it just feels magical.\\
It is my favorite Disney movie.\\
What's your favorite movie in general?\\
I think my favorite is The Sound of Music.\\
Really? Other than cartoons and stuff I can never get into musicals.\\
I love musicals. I really liked Phantom of the Opera.
\end{tcolorbox}


\section{Comparison with two-speaker data}
\label{sec:corpus_discussion}

\textit{How do self-dialogues compare with two-speaker conversations?}

To investigate the differences between our corpus and standard two-party conversations we adapted the VisDial framework \cite{DBLP:journals/corr/DasKGSYMPB16} to prompt two Workers to hold a conversation with one another. We collected 618 such conversations. The framework was modified by replacing the image with a set of popular topics from which the workers could choose: film, baseball, football, or fashion. The turn-based system was kept to enforce each party to wait for the other's reply as well as ensuring equal participation. We used the same criteria as for the single speaker collection with the modification that each worker had to send 15 messages before ending their task and receive US \$0.70. In cases where one Worker disconnected mid-task, we instructed the remaining Worker to imagine how the conversation would continue and finish their 15 messages accordingly in order to receive the payment. We label these conversations ``partially'' complete.

We found the self-dialogue setting to present multiple benefits over the standard 2-speaker approach. First, it is far simpler to set up, since the standard AMT interface can be used: coupling with a server to establish a connection between speakers is not required. Secondly, the setup of self-dialogues proved more efficient and convenient for the Worker. In the two-speaker setting, each participant must wait for the other party's reply.  This led to a median time for a Worker to complete a HIT of roughly 14.9 minutes (average response time 37 seconds). This proved to be unbearable for some of the workers and as a result the percentage of complete HITs was only 50.80\%. In contrast, the median completion time for the self-dialogues was 6.5 minutes.

Out of the 618 collected dialogues, a large portion was formed of either incomplete (31.71\%) or partial submissions (34.95\%). In the end the two speaker data contained a large number of self-dialogues with the added difficulty that the remaining Worker had to make sure the continuation was coherent.

\begin{figure}[htpb]

  \centering
  \includegraphics[trim={1.5cm 0.5cm 2.5cm 2cm},clip,width=1.0\columnwidth]{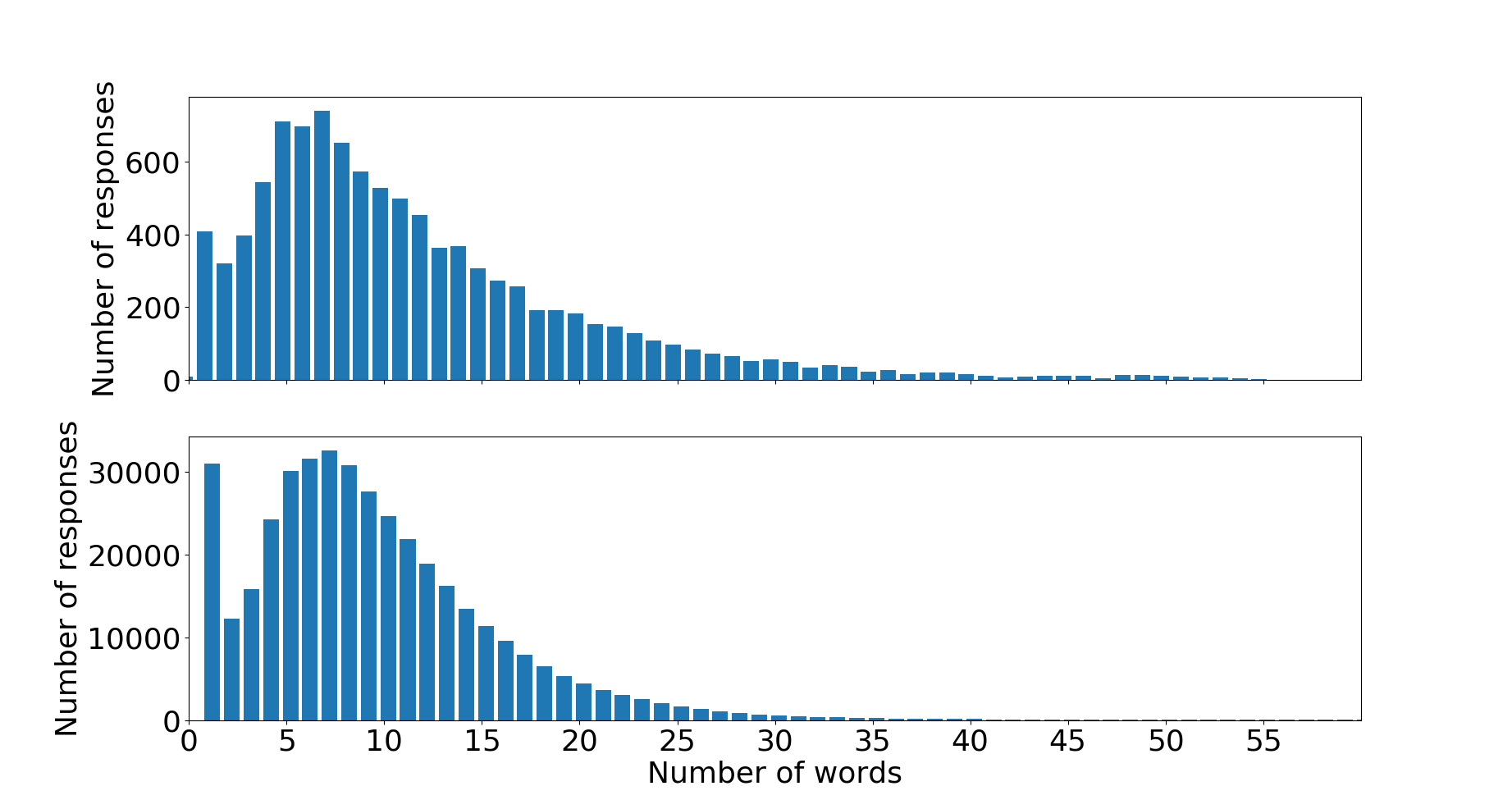}
  \caption{Histograms of response lengths from the two-speaker data (above) and self-dialogues (below).}
  \label{fig:lengths}
\end{figure}

Figure~\ref{fig:lengths} compares the length of responses between the two-speaker data and the self-dialogues. These resemble log-normal distributions and the long tails suggests there are some situations where  Workers engaged in very lengthy descriptions. However, the majority of the replies contain 6-7 words, apart from a large number of one-word responses (``hi'', ``yes'', etc.).

A closer inspection of the two types of data collected also show that the overall quality of the self-dialogue data exceeds that of the two-speaker conversations. The majority of the two-speaker dialogues contained either situations in which many clarifications from one of the participants were required or where one of the participants would not be particularly interested in the chosen topic and did not have any useful input. We have included a small set of samples from the two-speaker conversations for comparison and to illustrate the issues described in this section as well as the expected upper bound in quality\footnote{See Appendix~B for more examples.}.

\section{Comparison with OpenSubtitles}\label{sec:experiments}

\textit{How do self-dialogues compare to other available corpora?}

In order to further assess how well our data dataset fits the task we build a retrieval based conversational agent that given a query returns its most likely response. We analyze \textit{qualitatively} how well responses fetched from the self-dialogues pool are well fit to different types of queries by comparing with the OpenSubtitles corpus.

Our retrieval agent measures the distance between conversational queries $q$ with all conversational responses $r_i$, where $i$ is the response index. The closest match is found, and the next response, $r_{i+1}$, corresponds to the next line of the conversation. This is given as the agent's response. Response vectors $r_i$ and query vectors $q$ are given by a bag-of-words representation, with each word score given by the Inverse Document Frequency (IDF) score of that word. The IDF score of a word is given by the log of the total number of responses divided by the word's total frequency. Given conversational query $q$ and response vector $r_i$, the score of response $r_{i+1}$ is

\begin{equation}
    S(q,r_{i+1}) = \frac{{q^T r_i}}{\sqrt{r^T r}}.	
\end{equation}

Our first set of queries are designed to be more general, whereas our second set of queries are specific to the topics of data that we collected. We observed that whereas sometimes OpenSubtitles does return a well fitted response (see (\ref{retrex}.\ref{one})), it often fails when named entities are mentioned in the conversation (see \ref{retrex}.\ref{two} and \ref{retrex}.\ref{three})\footnote{See Appendix~C for more examples.}. OS denotes a response from OpenSubtitles and SD a response from the self-dialogues corpus.

\enumsentence{\label{retrex}
\begin{enumerate}
\item \label{one} \textit{how was your day}\\
OS: fantastic\\
SD: it is going good i have just been listening to music all day
\item \label{two} \textit{which harry potter movie did you like best}\\
OS: good one goyle\\
SD: i loved all of them
\item \label{three} \textit{who will win the superbowl}\\
OS: give me another magazine quick\\
SD: indianapolis colts
\end{enumerate}
}

\section{Conclusion}
In this work, we have proposed a novel approach to gathering data for open-domain conversation agents: self-dialogues through crowd sourcing, where a person is asked to create a fictitious two-party conversation. Analyses of the corpus have shown that self-dialogues present some advantages over standard two-party conversations in terms of cost and quality. We also compare a retrieval agent on OpenSubtitles and find the corpus presented here promising for comparisons when named entities are involved.

\bibliography{naaclhlt2018}
\bibliographystyle{acl_natbib}

\appendix
\section{Example self-dialogues}
\label{sec:example_dialogues}
We present unedited example dialogues from a range of the topics listed in Table~\ref{tab:data}.
\subsection{Movies}
\begin{itemize}
\itemsep0em
\item[\ding{172}] What is your absolute favorite movie?
\item[\ding{173}] I think Beauty and the Beast is my favorite.
\item[\ding{172}] The new one?
\item[\ding{173}] No, the cartoon.  Something about it just feels magical.
\item[\ding{172}] It is my favorite Disney movie.
\item[\ding{173}] What's your favorite movie in general?
\item[\ding{172}] I think my favorite is The Sound of Music.
\item[\ding{173}] Really?  Other than cartoons and stuff I can never get into musicals.
\item[\ding{172}] I love musicals.  I really liked Phantom of the Opera.
\end{itemize}

\subsection{Music}\label{sec:ed_sheeran}
\begin{itemize}
\itemsep0em
\item[\ding{172}] What do you think of Ed Sheeran?
\item[\ding{173}] I feel like he might be too overrated.
\item[\ding{172}] Are you kidding?? His voice is amazing!
\item[\ding{173}] Well I feel like the songs they place of his on the radio are catchy, but just not that great quality wise.
\item[\ding{172}] Well you definitely have to listen to his other music, because I would agree the songs on the radio are more overrated than they need to be.
\item[\ding{173}] Well what songs should I listen to then?
\item[\ding{172}] I would suggest listening to his new album, divide.
\item[\ding{173}] I've heard Shape of You.
\item[\ding{172}] But there are so many other amazing songs on the album!
\item[\ding{173}] Like what?
\item[\ding{172}] Well there's Galway Girl and Nancy Mulligan for starters.
\item[\ding{173}] I haven't even heard of those.
\item[\ding{172}] Exactly and those are the even better ones!
\item[\ding{173}] Which song is your favorite from the new album.
\item[\ding{172}] Well Supermarket Flowers really is so sweet, but I'd have to say Perfect is my favorite.
\item[\ding{173}] Is that a really pop song?
\item[\ding{172}] Not at all, because it's slower.  It feels more meaningful and powerful.
\item[\ding{173}] I'll have to give it a try, can I borrow your cd?
\item[\ding{172}] Of course! I'll bring it to work tomorrow!
\item[\ding{173}] Awesome, I'm excited to hear a different side of Ed Sheeran.
\end{itemize}

\subsection{NFL Football}
\begin{itemize}
\itemsep0em
\item[\ding{172}] What NFL team will have the best regular season record this year?
\item[\ding{173}] Hard not to go with the Patriots, right?
\item[\ding{172}] I'm not so sure, this year.
\item[\ding{173}] Why not?
\item[\ding{172}] Well, for one thing, I think the Dolphins might be a good team this year.
\item[\ding{173}] If their division gets more challenging, that will make it tougher for them to roll through the regular season, true.
\item[\ding{172}] Also, at SOME point Tom Brady's age will catch up to him.
\item[\ding{173}] Hasn't happened yet! The dude just won the Super Bowl.
\item[\ding{172}] But eventually, it will happen. Maybe through injuries.
\item[\ding{173}] Eventually, you'll be right about Brady, but I wouldn't bet on it this year.
\item[\ding{172}] I think the Seahawks have a really good chance to have the best record.
\item[\ding{173}] Hmm, I think they'll be good, but why do you say best record?
\item[\ding{172}] First of all, they have the best home field advantage in the league.
\item[\ding{173}] Yep, no better home field advantage.
\item[\ding{172}] Second, I think their division is soft.
\item[\ding{173}] You're not a believer in the Cardinals?
\item[\ding{172}] I'm not sold on their QB situation, and I think both the Rams and the 49ers will be bad.
\item[\ding{173}] If all three teams are mediocre to bad, then Seattle should have a great record.
\item[\ding{172}] Yeah, that's just how I see it falling out this year.
\item[\ding{173}] I hope it's not another Seattle/New England Super Bowl. I want some new talent!
\end{itemize}

\subsection{Lady Gaga}\label{sec:lady_gaga}
\begin{itemize}
\itemsep0em
\item[\ding{172}] What is your favorite Lady GAGA album?
\item[\ding{173}] I loved the Born this Way album.  It had so many great hits on it.
\item[\ding{172}] Which Lady GAGA video is your favorite?
\item[\ding{173}] The Telephone video was great.  It was like a mini-movie.
\item[\ding{172}] Do you prefer the older Lady GAGA music or the more recent singles?
\item[\ding{173}] I much prefer the older singles like Poker Face and Paparazzi.  They had a more upbeat feel.
\item[\ding{172}] Which Lady GAGA tour was your favorite?
\item[\ding{173}] The Monster Ball Tour was amazing.  She performed all her greatest hits and she was very inspiring.
\item[\ding{172}] How do you think the music in the upcoming movie A Star is Born will compare to her traditonal work?
\item[\ding{173}] I think Lady GAGA will do an amazing job in the movie.  I'm really looking forward to how she will recreate the role and music for the origianl film.
\end{itemize}
\subsection{Transition music/movies}
\label{sec:transition_music_movies}
\begin{itemize}
\itemsep0em
\item[\ding{172}] I found an actual snake in my husband's boot.
\item[\ding{173}] That is like something out of a movie. Speaking of, have you watched any good westerns?
\end{itemize}

\section{Example two-speaker conversations}
\label{sec:two_person_conversations}
We present a small set of dialogues from the two speaker conversations. If a message is missing or the dialogue is short, then the worker(s) left the conversation.

\subsection{Corrections}
\begin{itemize}
\itemsep0em
\item[\ding{172}] who is your favorite nfl football team?
\item[\ding{173}] packers, you?
\item[\ding{172}] steelers
\item[\ding{173}] Ha, I'm talking to another Steelers fan now.  You gonna tell me Rodgers has fallen off too?
\item[\ding{172}] we have an in-house division rivalry, my daughter likes the titans.
\item[\ding{173}] oh yeah?  That's cool.  Ive been to a game in Nashville.  lots of fun
\item[\ding{172}] i've been to a game in cincinnati as well as pittsburgh.
\item[\ding{173}] Oh yeah?  Ive never even been to the cities.  I had a layover in Philly, thats about as close as I've gotten
\item[\ding{172}] my brother lives in bridgeville, just outside pittsburgh ... i've often thought about relocating just to be closer to my boys LOL
\item[\ding{173}] Well yeah.  Packers and Steelers are very similar.  Rich history
\item[\ding{172}] agreed....but steelers still have the most rings, thus far anyway lol
\item[\ding{173}] But the Packers have the most Championships, and that's not getting caught
\item[\ding{172}] are you referring to division championships?
\item[\ding{173}] NFL Championships
\item[\ding{172}] steelers have 6 ... nobody else has 6, yet
\item[\ding{173}] 13 World Championships for the Packers, 6 for the Steelers, look it up
\item[\ding{172}] they've played 13 times, but have only won 4
\item[\ding{173}] No.  Before the Super Bowl
\item[\ding{172}] prior to 1933?
\item[\ding{173}] I cant copy and paste links.  Google packers world championships and click on the first link
\item[\ding{172}] i just saw the info about world championships
\item[\ding{173}] Bingo.  I mean it;s deadball era, but ill take it, haha
\item[\ding{172}] lol ... i suppose just like myself,  a die-hard fan...nothing wrong with that!
\item[\ding{173}] Exactly, hmm, what happened last time we met in the Super Bowl? 
\item[\ding{172}] lol yet again, another point proven, and taken!
\item[\ding{173}] Well good luck this year, Love Ben
\item[\ding{172}]  i think he's a good quarterback.. but still believe the quarterback is only as good as his offensive line
\item[\ding{173}] Very very smart.  But when you have Ben, AB, and Bell, they do kinda make up for it 
\item[\ding{172}] agreed....you'd think the steelers would have another ring or two with that trio
\item[\ding{173}] Bingo
\end{itemize}

\subsection{Conversation dying off}
\begin{itemize}
\itemsep0em
\item[\ding{172}] hello
\item[\ding{173}] Hey , what do you think of the act of terror in Manchester?
\item[\ding{172}] i think it sucks..
\item[\ding{173}] yea really sad all those children killed
\item[\ding{172}] the world is getting selfish
\item[\ding{173}] yea it's definetly more divided 
\item[\ding{172}] its going downhill at an even pace..
\item[\ding{173}] Well the US needs to protect itself 
\item[\ding{172}] i agree, fav football team?
\item[\ding{173}] I like the vikings  how about you 
\item[\ding{172}] steelers all the way. but ben is getting old
\item[\ding{173}] yea he is a tough dude  takes a ton of hits but it probably is working against him now 
\item[\ding{172}] 2 more yrs tops, and he is done, sadly
\item[\ding{173}] well he had a great career 
\item[\ding{172}] a fairy tale career for sure.
\item[\ding{173}] How about baseball?
\item[\ding{172}] im in Ga. I'M A BRAVES FAN, BUT NOT FALCONS...LOL... YOU?
\item[\ding{173}] I'm in New Jersey  and I'm a Twins fan well  Minnesota fan in every sport 
\item[\ding{172}] TWINS ARE A GREAT BALL TEAM
\item[\ding{173}] Well there having a good start this year .  Unfortunately they always run into the Yankees in the playoffs
\item[\ding{172}] YANKEES ARE TOUGH... WE HAD OUR DAY WITH BOBBY COX, NOW WE SUCK
\item[\ding{173}] The braves and Twins played a classic world series in 1991   one of the best ever
\item[\ding{172}] I REMEMBER
\item[\ding{173}] How is the weather going to be for the Holiday weekend?
\item[\ding{172}] rain, ughh and u?
\item[\ding{173}] -
\item[\ding{172}] ok
\item[\ding{173}] -
\item[\ding{172}] ok
\item[\ding{173}] -
\end{itemize}

\subsection{Failed continuation}
\begin{itemize}
\itemsep0em
\item[\ding{172}] so how about those Steelers
\item[\ding{173}] Ravens fan
\item[\ding{172}] oh boy, im sorry for you..
\item[\ding{173}] Me too. They've been mediocre ever since the superbowl win
\item[\ding{172}] our nemisis
\item[\ding{173}] Big Ben is getting old though
\item[\ding{172}] it always seems to be a good game when they play each other though
\item[\ding{173}] If you like defense, definitely
\item[\ding{172}] i suspect he has 2 more seasons at best.
\item[\ding{173}] Yeah. I think so too. And then what?
\item[\ding{172}] then , Honestly I DONT KNOW, THEY HAVE BEEN PROSPECTING but nothing good yet..
\item[\ding{173}] -
\item[\ding{172}] they will get someone
\item[\ding{173}] -
\item[\ding{172}] then it will need work
\item[\ding{173}] -
\item[\ding{172}] i still love them
\item[\ding{173}] -
\item[\ding{172}] they are my home team
\item[\ding{173}] -
\item[\ding{172}] they are skilled
\item[\ding{173}] -
\item[\ding{172}]the best
\item[\ding{173}] -
\item[\ding{172}] and play well
\item[\ding{173}] -
\item[\ding{172}] im confident
\item[\ding{173}] -
\item[\ding{172}] they will win superbowl this yr
\item[\ding{173}] -
\item[\ding{172}] hello
\item[\ding{173}] -
\end{itemize}

\subsection{Short acceptable but not very interesting conversation}
\begin{itemize}
\itemsep0em
\item[\ding{172}] Do you like fashion?
\item[\ding{173}] I don't like fashion. 
\item[\ding{172}] whats your favorite movie 
\item[\ding{173}] My favorite movie is idiocrazy. 
\item[\ding{172}] Ive never seen it. what genre is it? 
\item[\ding{173}] it is a political satire movie. 
\item[\ding{172}] ahh ok. is satire your favorite genre? 
\item[\ding{173}] yes, satire is funny so it is my favorite. 
\item[\ding{172}] I prefer drama and true crime 
\item[\ding{173}] Why do you like drama?
\item[\ding{172}] I like the more true to life storylines 
\item[\ding{173}] Is that also why you like true crime? 
\item[\ding{172}] Yes, I took criminal justice in school. 
\item[\ding{173}] what makes true to life storylines good for movies? 
\item[\ding{172}] I just feel like i get more engrossed. why do you like comedy?
\item[\ding{173}] Comedy is funny, i'm a simple person that can't understand much else. 
\item[\ding{172}] SO do you like more slapstick comedy? or spoof films
\item[\ding{173}] I like all sorts of comedic films, I find all various types of comedy to be funny. I'm easily amused.
\item[\ding{172}] I like comedy when i've had a bad day. easy to take your mind off life 
\item[\ding{173}] Do you think your education has had an impact on how you view film?
\item[\ding{172}] I think so.  It made me look at how people think so true crime lets me try to determine why they did what they did 
\item[\ding{173}] Who is your favorite actor?
\item[\ding{172}] tom hiddleston. you?
\item[\ding{173}] Robert De Niro, why is tom your favorite?
\item[\ding{172}] I like his personality. Plus hes so cute. Why do you like DeNiro?
\item[\ding{173}] Because he was a major part of the movies I enjoyed learning about in school.
\item[\ding{172}] That makes sense. whats your favorite Deniro movie? 
\item[\ding{173}] Mean streets, have you ever seen it?
\item[\ding{172}] I dont think I have 
\item[\ding{173}] It is a good movie.
\end{itemize}

\subsection{Very good conversation}
\begin{itemize}
\itemsep0em
\item[\ding{172}] Hi. Logan was a great movie. 
\item[\ding{173}] Well, it looks like the new Baywatch movie was a total flop.
\item[\ding{172}] It didn't know there was a new Baywatch movie. 
\item[\ding{173}] Yeah, it stars The Rock. But they added lots and lots of guns and made it very un-Baywatchy.
\item[\ding{172}] The Rock normally does good stuff, but you can't have Baywatch without the Hoff. 
\item[\ding{173}] So true. I want to see Logan. Can't believe I've not seen it yet.
\item[\ding{172}] Logan is a relentlessly brutal film, and very emotional. I was actually tearing up at the ending. 
\item[\ding{173}] Oh wow, cool.  I've about reached my limit on super hero movies, but I definitely want to see this one. Do you watch many of the super hero franchise movies?
\item[\ding{172}] Dr Strange is the last one I've seen.   But I do try to keep up with Marvel's films. 
\item[\ding{173}] Do you go to the theater or do you stream like on Netflix or Apple TV or something like that?
\item[\ding{172}] I stream, mostly. I avoid the theater. 
\item[\ding{173}] Same here. Netflix is just unbelievable right now. My go-to places are Netflix, Hulu, and Amazon originals.  So much quality there.
\item[\ding{172}] I need to watch the Amazon shows.  I have Prime for the shipping, but haven't really watched their videos. 
\item[\ding{173}] Oh wow, you're in for a treat. Great original programming, too.  I just wrapped up the 3rd season of Bosch, which is an Amazon original. Sort of a modern Noir detective show. They have bunch of older HBO and Showtime stuff on Amazon Prime, too. So definit
\item[\ding{172}] I will. The great thing about streaming services is that they can take a lot more risks than traditional networks. 
\item[\ding{173}] Yeah, definitely. So many of the network shows just feeling like they've passed through the corporate network executive filter. Do you "binge" watch shows?
\item[\ding{172}] All the time. Mostly old 90s shows. 
\item[\ding{173}] What are the good 90s shows?
\item[\ding{172}] It depends on what your personal taste are.  I like the syndicated adventure shows, but that might just be nostalgia talking.  Highlander: The Series was one of my favorites.  Babylon 5 is also good, though it starts off really slow, and would have been c
\item[\ding{173}] Oh wow, I've forgotten about those.  Those were Syfy Network, right? I need to remember about that channel. Where you a fan of The Doom? Can't remember if that was Syfy on Amazon.
\item[\ding{172}] Highlander was originally in First Run Syndication.  USA had reruns, then Syfy had some, too.  Babylon 5 was on PTEN, which was Warner Brother's attempt to start their own network, but ended up being just a syndication package.  It moved to TNT for the fi
\item[\ding{173}] Complicated route for both those shows. Oh, I mean The Dome.  Hahah.  It might be an Amazon Original. An invisible dome encapsulates (?) this small town, trapping everyone on the inside. Fantastic first two season. Got a bit off the rails after that. My w
\item[\ding{172}] It was on CBS, and was based on a book by Stephen King.  I've never watched it, I might check it out. 
\item[\ding{173}] That's right. It was Stephen King. My daughter (17) recently watched Stand By Me. She couldn't believe that was a Stephen King story.  In fact, most of my favorite King movies aren't his horror stories.
\item[\ding{172}] Yeah. His horror stories tend to be predictable, it's the non-horror stuff that gets interesting. 
\item[\ding{173}] Was  King's The Stand a movie or mini-series of some sort? Or am I just imagining images from the book? 
\item[\ding{172}] There was a mini-series, on ABC I think. 
\item[\ding{173}] Oh on ABC. Probably not so good.  That show is not network adaptable. 
\end{itemize}

\section{Retrieval bot responses}\label{sec:exp}
We present preliminary experimental results with a retrieval bot evaluated on a set of predefined queries. Q denotes a query, OS a response from OpenSubtitles and SD a response from the self-dialogues presented in this paper. See Section~\ref{sec:experiments} for details.
\subsection{General queries}
\begin{itemize}
\itemsep0em
\item[Q] \emph{hello}
\item[OS] what a surprise
\item[SD] hi there
\end{itemize}
\begin{itemize}
\itemsep0em
\item[Q] \emph{how are you}
\item[OS] why is your pajama to on
\item[SD] doing good and you
\end{itemize}
\begin{itemize}
\itemsep0em
\item[Q] \emph{what do you want to talk about}
\item[OS] why did you invite me to stay for the holidays
\item[SD] let's talk about football
\end{itemize}
\begin{itemize}
\itemsep0em
\item[Q] \emph{how was your day}
\item[OS] fantastic
\item[SD] it is going good i have just been listening to  music all day
\end{itemize}

\subsection{Topical queries}
\begin{itemize}
\itemsep0em
\item[Q] \emph{what is your favorite movie}
\item[OS] star wars of course
\item[SD] oh that's a tough one i think the prestige
\end{itemize}
\begin{itemize}
\itemsep0em
\item[Q] \emph{which harry potter movie did you like best}
\item[OS] good one goyle
\item[SD] i loved all of them
\end{itemize}
\begin{itemize}
\itemsep0em
\item[Q] \emph{what is your favorite band}
\item[OS] elvis costello
\item[SD] i like acdc
\end{itemize}
\begin{itemize}
\itemsep0em
\item[Q] \emph{who will win the superbowl}
\item[OS] give me another magazine quick
\item[SD] indianapolis colts
\end{itemize}

\end{document}